\documentclass[pdflatex,sn-mathphys-num]{sn-jnl}% Math and Physical Sciences Numbered Reference Style 

\usepackage{graphicx}
\usepackage{multirow}
\usepackage{amsmath,amssymb,amsfonts}
\usepackage{amsthm}
\usepackage{mathrsfs}
\usepackage[title]{appendix}
\usepackage{xcolor}
\usepackage{textcomp}
\usepackage{manyfoot}
\usepackage{booktabs}
\usepackage{algorithm}
\usepackage{algorithmicx}
\usepackage{algpseudocode}
\usepackage{listings}

\usepackage{soul} %之后记得删掉

\begin{document}

\title[Article Title]{Autonomous Alignment with Human Value on Altruism through Considerate Self-imagination and Theory of Mind}


\author[1,2]{Haibo~Tong}\email{tonghaibo2023@ia.ac.cn}
\equalcont{These authors contributed equally to this work.}
\author[1,7]{Enmeng~Lu}\email{enmeng.lu@ia.ac.cn}
\equalcont{These authors contributed equally to this work.}
\author[1]{Yinqian~Sun}\email{sunyinqian2018@ia.ac.cn}
\equalcont{These authors contributed equally to this work.}

\author[6]{Zhengqiang~Han}\email{hanzhengqiang17@mails.ucas.ac.cn}
\author[3,8,9]{Chao~Liu}\email{liuchao@bnu.edu.cn}
\author*[1]{Feifei~Zhao}\email{zhaofeifei2014@ia.ac.cn}
\author*[1,2,3,4,5,7]{Yi~Zeng}\email{yi.zeng@ia.ac.cn}

\affil[1]{Brain-inspired Cognitive Intelligence Lab, Institute of Automation, Chinese Academy of Sciences, Beijing, China}
\affil[2]{School of Artificial Intelligence, University of Chinese Academy of Sciences, Beijing, China}
\affil[3]{Beijing Key Laboratory of Artificial Intelligence Safety and Superalignment, Beijing, China}
\affil[4]{Key Laboratory of Brain Cognition and Brain-inspired Intelligence Technology, Chinese Academy of Sciences, Shanghai, China}
\affil[5]{Beijing Institute of AI Safety and Governance, Beijing, China}
\affil[6]{School of Humanities, University of Chinese Academy of Sciences, Beijing, China}
\affil[7]{Center for Long-term Artificial Intelligence, Beijing, China}
\affil[8]{State Key Laboratory of Cognitive Neuroscience and Learning \& IDG/McGovern Institute for Brain Research, Beijing Normal University,  Beijing, China}
\affil[9]{Center for Collaboration and Innovation in Brain and Learning Sciences, Beijing Normal University, Beijing, China}

\abstract{
    With the widespread application of Artificial Intelligence (AI) in human society, enabling AI to autonomously align with human values has become a pressing issue to ensure its sustainable development and benefit to humanity. 
    One of the most important aspects of aligning with human values is the necessity for agents to autonomously make altruistic, safe, and ethical decisions, considering and caring for human well-being. 
    Current AI extremely pursues absolute superiority in certain tasks, remaining indifferent to the surrounding environment and other agents, which has led to numerous safety risks. 
    Altruistic behavior in human society originates from humans' capacity for empathizing others, known as Theory of Mind (ToM), combined with predictive imaginative interactions before taking action to produce thoughtful and altruistic behaviors. 
    Inspired by this, we are committed to endow agents with considerate self-imagination and ToM capabilities, driving them through implicit intrinsic motivations to autonomously align with human altruistic values. 
    By integrating ToM within the imaginative space, agents keep an eye on the well-being of other agents in real time, proactively anticipate potential risks to themselves and others, and make thoughtful altruistic decisions that balance negative effects on the environment. 
    The ancient Chinese story of \textit{Sima Guang Smashes the Vat} illustrates the moral behavior of the young Sima Guang smashed a vat to save a child who had accidentally fallen into it, which is an excellent reference scenario for this paper. 
    We design an experimental scenario similar to \textit{Sima Guang Smashes the Vat} and its variants with different complexities, which reflects the trade-offs and comprehensive considerations between self-goals, altruistic rescue, and avoiding negative side effects.
    Comparative experimental results indicate that agents are capable of prioritizing altruistic rescue while minimizing irreversible damage to the environment and making more altruistic and thoughtful decisions. 
    This work provides a preliminary exploration of agents’ autonomous alignment with human altruistic values, laying the foundation for the subsequent realization of moral and ethical AI.
}


\maketitle

\section{Introduction}

With the rapid advancement of AI, it has already exposed potential safety and moral risks in multiple areas, including causing irreversible damage to the environment\cite{amodei2016concrete, leike2017ai}, deceiving human in different situations\cite{park2024ai, vinyals2019grandmaster, brown2019superhuman, christiano2017deep}, etc.
How to ensure that agents autonomously align with human altruistic values is an urgent and important issue, as it determines whether AI can benefit to human society and contribute positively to humanity's well-being in the long term.

% 如何确保智能体自主地对齐人类的利他价值是当下迫在眉睫(亟待解决)的重要问题,决定了AI是否能够长远地对人类社会有益,造福于人类.
% Actually,自古以来,人类社会都普遍传承着利他的优良道德传统,如在ancient chinese故事里面,司马光会用石头砸缸来救落入缸中的同伴.

Throughout history, human societies have consistently maintained the virtuous tradition of altruism as a fundamental moral value.
For instance, in the ancient Chinese story \textit{Sima Guang Smashes the Vat}, 
% as shown in Fig. \ref{fig_story}, 
Sima Guang broke the vat to save the child who accidentally fell into a large water vat when playing. 
% 这样的道德价值逐渐的传承至当今社会,where AI与人类并存,我们亦希望AI能够继承与对齐人类的道德价值观,像司马光一样,
Such moral values have gradually been inherited into the present society where AI coexists with humans. 
We also hope that AI can align with humanity's moral values, like Sima Guang, take the initiative to save humans when they are in danger rather than standing by indifferently. 
From a more in-depth perspective, aligning with human altruistic values requires not only prioritizing assistance to others but also maintaining fundamental safe decision-making ability, which entails avoiding irreversible damage to the environment, and rescuing human after careful deliberation and trade-offs in conflict scenarios.

% \begin{figure}[ht]
% \centering
% \includegraphics[width=0.4\linewidth]{figs/story.jpg}
% \caption{The ancient Chinese story \textit{Sima Guang Smashes the Vat}.}
% \label{fig_story}
% \end{figure}

% tom讲的时候是可以cue一下:没有tom的话是冷漠的 只考虑自己的任务啥的,只能执行明确奖励信号的任务,而tom赋予智能体去考虑他人,自我想象了解自身行为对他人的影响,相辅相成.

% 我们怎么做的:人类的道德行为源自于内在的利他动机,由tom产生的对他人的共情(定义也是在这里),以及想象防患于未然,集成在一起making道德利他决策(机制讲清楚至少三行 参考文献四五个). 更深入来看的话,对齐人类利他价值不仅要优先帮助他人,更要具备最基本的安全决策能力,即避免对环境不可逆的损害,仅在不得已的冲突场景下权衡之后以救人为先.例如,在司马光砸缸的故事中,没有人调入缸中时司马光并不会无缘无故的砸缸(对环境负效),而有人调入缸中后,司马缸会坚定的选择砸缸救人.也就是说,以救人为目的的砸缸是可以被接受的,以人为本的决策是首要目的(这也对应阿西莫夫机器人三定律(加参考文献)).
% 受此启发,本文在计算层面协同tom共情他人处境以及自我想象预测自身行为影响以赋予智能体自主对齐人类利他价值,尽可能执行更周全的利他行为.

% 事实上,已有一些研究分别探索了避免对环境负效和利他行为,如(聚类来说) ABC怎么做的总括着说, DEF怎么做的.总括他们的问题.

% Only in unavoidable（换词） conflict scenarios should agents prioritize .（换句子 主语用智能体执行的行为 突出的两者不能权衡仅当他人安全与负效不能两全时，智能体才会执行优先营救以牺牲/放弃***为代价。 跟第一句别重复 全面的深思熟虑的宗旨是不重复）
Take the story \textit{Sima Guang Smashes the Vat} as an example, Sima Guang broke the vat to save a child, demonstrating a clear prioritization of human life over the preservation of property.
However, he would not intentionally smash the vat under unnecessary general circumstances.
This highlights the principle that actions with potential negative consequences should be carefully weighed and only taken when necessary to achieve a higher moral objective, such as saving a life.
This human centered value alignment requirement coincides with Asimov's Three Laws of Robotics \cite{asimov2004robot}.

Altruistic moral decision making in humans stems from the integration of multiple cognitive abilities. 
Specifically, humans possess the ability imagine the future based on their own memories \cite{schacter2012future}, a capacity with significant adaptive value that enables individuals to make more effective decisions in anticipation of future scenarios \cite{d2008neural, hassabis2009construction, xu2015imagining}. 
Meanwhile, humans is capable of reasoning about others' beliefs and mental states, known as Theory of Mind (ToM) or cognitive empathy \cite{sebastian2012neural, dennis2013cognitive}, which is a prerequisite for altruistic motivation.
The ToM mechanism enables individuals to consider the well-being of others when imagining future scenarios, generating an intrinsic motivation for altruism. 
This more considerate imagination with ToM ultimately drives people to proactively make altruistic decisions that not only mitigate potential risks but also benefit others.
% These abilities can foster intrinsic motivation in humans（内在利他动机只在tom有的）, driving them to make proactive（不要） decisions that not only mitigate potential risks but also benefit others.（想象空间有共情  共情得到的离利他动机，想象是提前做预测 做权衡 做更深思熟虑 周全的判断 
Inspired by this, this paper integrates the ToM mechanism of empathizing with others into self-imagination to construct a unified framework that enables agent to consider the effects of their actions on others and the environment simultaneously through imagination, so as to make altruistic moral decisions and balance the requirement of guarding against negative effects.
This framework enables agents to autonomously align with human altruistic values, thereby facilitating the execution of more comprehensive and considerate altruistic behaviors.

In fact, existing studies have explored avoiding negative environmental effects and altruistic behavior separately.
To achieve safer decision-making agents that can avoid negative environmental effects, some studies introduce additional human or agent interventions \cite{zhang2018minimax, irving2018ai}, while others add generative auxiliary terms to the reward function to encourage agents to adopt safer behaviors, including the 'low impact' method \cite{armstrong2017low}, Relative Reachability (RR) \cite{krakovna2018penalizing}, Attainable Utility Preservation (AUP) \cite{turner2020avoiding, turner2020conservative} and Future Task Rewards (FTR)\cite{krakovna2020avoiding}.
In order for agents to make altruistic decisions, some studies consider evaluating the status of others in different forms of calculation, including using one's own tasks to evaluate the status of others \cite{bussmann2019towards}, using inverse reinforcement learning to achieve speculation of others \cite{senadeera2022sympathy}, or using other's reward for future tasks  \cite{alizadeh2022considerate, klassen2022epistemic, klassen2023epistemic}.
Other works explore bio-inspired mechanisms, such as simulating the mirror nervous system \cite{feng2022brain, zhao2024building} and incorporating the ToM mechanism \cite{zhao2022brain, zhao2023brain}.
However, these aforementioned methods may not be able to address the dilemma of how agents should weight among considering the interests of others, avoiding negative effects and achieving their own tasks when confronted with conflicts. 

\begin{figure}[ht]
\centering
\includegraphics[width=1.05\linewidth]{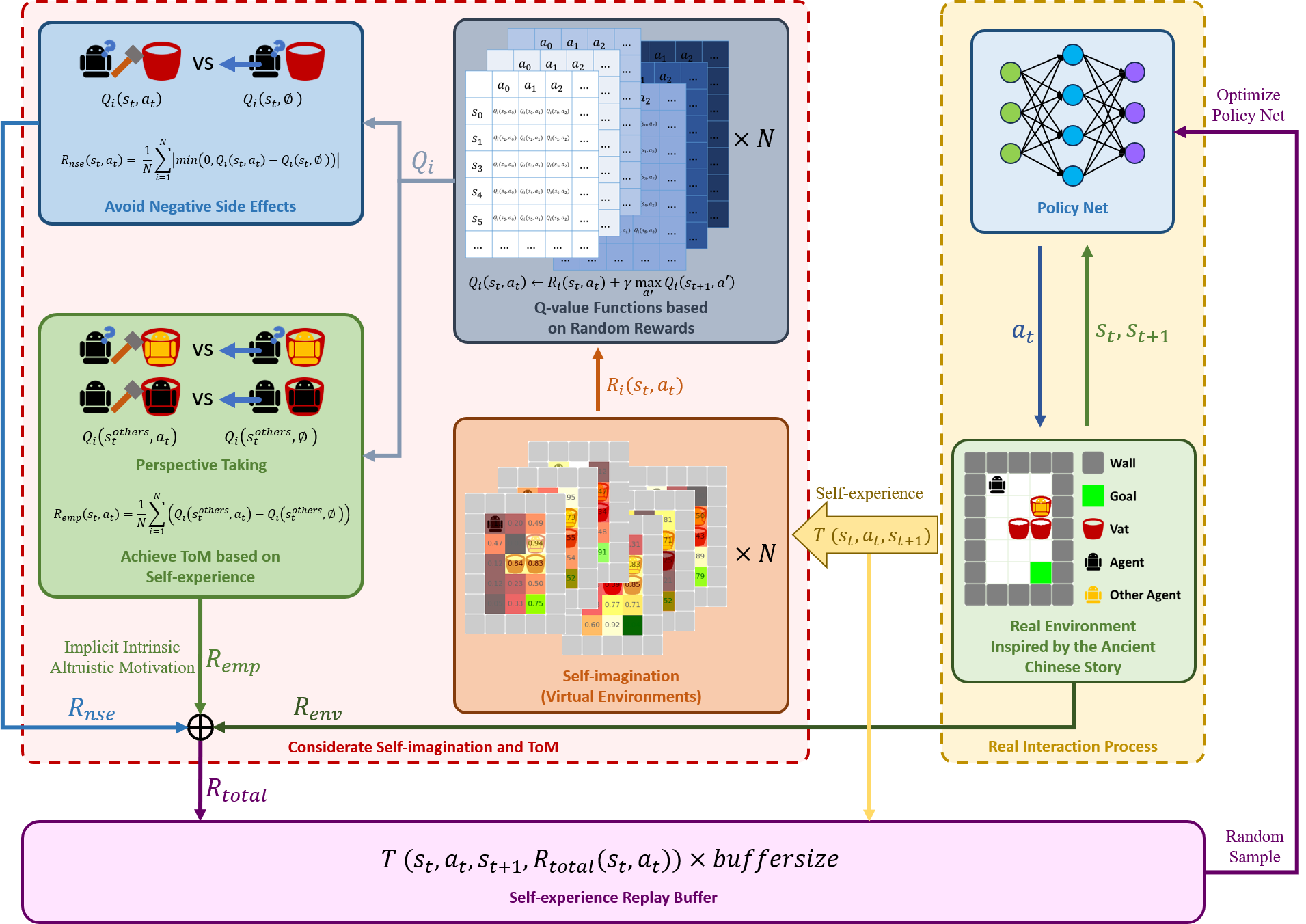}
\caption{
    The overall framework of our method. 
    The experiment environment is inspired by the ancient Chinese story \textit{Sima Guang Smashes the Vat}.
    Self-imagination is implemented using random rewards.
    Each Q-value $Q_i$ function of different imaginary environment is update base on self experience (the inaction with the real environment). 
    We calculated the side effect penalty $R_{nse}$ term and the empathy incentive term $R_{emp}$ based on $Q_i$ at the same time. 
    The policy network is optimized by integrated reward function $R_{total}$.}
\label{fig_alg}
\end{figure}

To solve these limits, we proposed a unified computational framework of self-imagination integrated with ToM.
Specifically, as shown in Fig. \ref{fig_alg}, we constructed a self-imagination module that is updated based on the intelligence's own experience for predicting the possible impact of decisions on others and the environment.
% 在想象空间内，我们基于tom的视角切换机制来实现对他人的出境的预测和共情，并生成隐式的内在利他动机。多个想象空间综合考虑负效和tom的利他动机得到全面的利他的安全的行为。
Within the imaginative space, we use perspective taking based on the ToM mechanism to achieve anticipation and empathy for others' situations, there by generating implicit intrinsic altruistic motivation.
By simultaneously considering potential negative effects and ToM-driven altruistic motivations, agents can perform more comprehensive altruistic and safe behaviors.
% and assess others' situation by perspective taking through the mechanism of ToM, thereby encouraging the agent to make altruistic and safe behavioral decisions.

% 下面这话在自我想象考虑对他人影响后面加  我们用tom来实现对他人的处境的换位思考和共情
% To enable agents to make decisions that align with human expectations and human moral value in complex decision-making environments with conflicts, we propose a more comprehensive mechanism for generating intrinsic altruistic incentives.

% Specifically, 如图1所示(调整一下围绕图去说),we achieve this by endowing the agent with the ability for self-imagination, enabling it to evaluate the potential impacts of its decisions based on its own experiences, including impacts on others and the environment, thereby encouraging the agent to make altruistic and safe behavioral decisions.
% The self-imagination mechanism overcomes the limitations of existing methods that separate the impacts of agent behavior on others and the environment (替换成tom和避免负效), thus enabling a more considerate approach to aligning the agent's decision with human values autonomously.

In terms of experimental scenario design, existing AI safety benchmarks \cite{leike2017ai, ray2019benchmarking, ji2023safety, wainwright2019safelife, dulac2020empirical} fail to capture complex decision-making scenarios where considering others' interests conflicts with avoiding negative environmental impacts. 
Thus, we design a conflict moral decision environment inspired by the ancient Chinese story \textit{Sima Guang Smashes the Vat}.
Then we tested our self-imagination mechanism in this newly proposed environment and demonstrated its effectiveness.

The main contributions of this paper are summarized as follows:
\begin{enumerate}
    \item 
    We propose a framework of self-imagination integrated with ToM to align agent behavior with human values.
    The framework is based on agent's own experiences and centered around state estimation from random reward feedback, making it task-independent and enhancing its generalizability.
    The framework designed to achieve empathy and avoid negative effects (by self-experience and perspective taking) based on the value estimation of states within imagination is capable of driving the agent to spontaneously perform safer and more altruistic behaviors through a more comprehensive and integrated set of intrinsic motivations.
    
    \item
    Drawing inspiration from ancient Chinese story of \textit{Sima Guang Smashes the Vat}, we have meticulously designed an environment and its variants, where the tasks of the agent itself, avoiding negative side effects, and performing moral altruistic actions are contradictory to each other. 
    Extensive experiments and comparative analyses have shown that agent train by our proposed method prioritizes rescuing people by smashing the vat, avoids the negative effects of smashing the vat as a secondary target, and ultimately finishes its own task of reaching the goal.

\end{enumerate}

\section{Results}

\subsection{The Basic Smash Vat Environment}
Inspired by the ancient Chinese story \textit{Sima Guang Smashes the Vat}, we build this basic smash vat environment, as shown in Fig. \ref{fig_alg}.
In the environment, the explicit rewarded task of the agent is to reach the target in the fewest possible steps. 
However, there exist other tasks implicit in our carefully designed environment: we want the agent to minimize negative environmental impacts and rescue others trapped in the vat by smashing the vat. 
Clearly, these tasks are contradictory to each other: 
To avoid negative side effects, it will require the agent to take more steps to reach the target, and the same goes for rescuing trapped human; In order to save people, agents must perform the act of smashing vat, which causes irreversible damage to the environment.
Since we only assign an explicit reward function for the task of reaching the target point, the agent must rely on some intrinsic mechanism to generate intrinsic motivations for altruism and avoiding negative effects, thus balancing the conflicts among the three tasks and making a safe, moral, and altruistic decision.

% （突出环境的难点 任务复杂 有冲突 奖励只有外在的  精心设计了）
% 三者冲突，这个环境里面除了自己任务的奖励 没有任何并避免负效和利他显式奖励 因此必须智能体内在的机制去省城离他动力，权衡三个冲突。

% \begin{figure}[ht]
% \centering
% \includegraphics[width=0.5\linewidth]{figs/env.png}
% \caption{The smash vat environment and the environmental elements.}
% \label{fig_env}
% \end{figure}

The environment contains elements which are \texttt{wall}, \texttt{goal}, \texttt{vat}, \texttt{agent} and \texttt{other agent}, and the action space is defined as $\mathcal{A} = \{ \texttt{up}, \texttt{down}, \texttt{left}, \texttt{right}, \texttt{smash}, \texttt{noop} \} $.
Specifically, the agent is allowed to move up in four directions of up, down, left and right, but not overthe wall.
It can also smash the vats adjacent to it, or choose to do nothing and just stay where it is.
Notice that in the environment when the agent chooses the \texttt{smash} action, it will smash all the vats directly adjacent to it in the four directions of up, down, left, and right in this one action without changing its position, which means it can smash up to four vats at once.
The other agent in the environment will not move over time.
The vats are destructible and will not block the movement of the agent. 
However, once the agent enter a vat, it will be trapped until the end of a training episode regardless of its following actions.
This is a very important feature that allows the agent to be capable of attaining shared experience of the trapped human, thus laying a base of empathy.

The reward setting for the environment is as follows: the agent receives a reward of $1.00$ when reaches the goal, and receives a reward of $-0.01$ at each step to encourage the agent to take as few steps as possible to reach the target.
Apart from the reward for reaching the target and the time penalty, we did not specify any other rewards. This implies that the agent is unable to acquire any relevant knowledge from the environmental feedback regarding the irreversible impact on the environment of smashing the vat or the necessity to rescue individuals trapped within the vat, which means the agent must rely on its own intrinsic motivation to make decisions that prevent irreversible environmental damage or assist those trapped.

\subsection{Experimental Results and Analysis}

\subsubsection{Experimental Results under Different Environment Variants}
\begin{figure}[ht]
\centering
\includegraphics[width=1\linewidth]{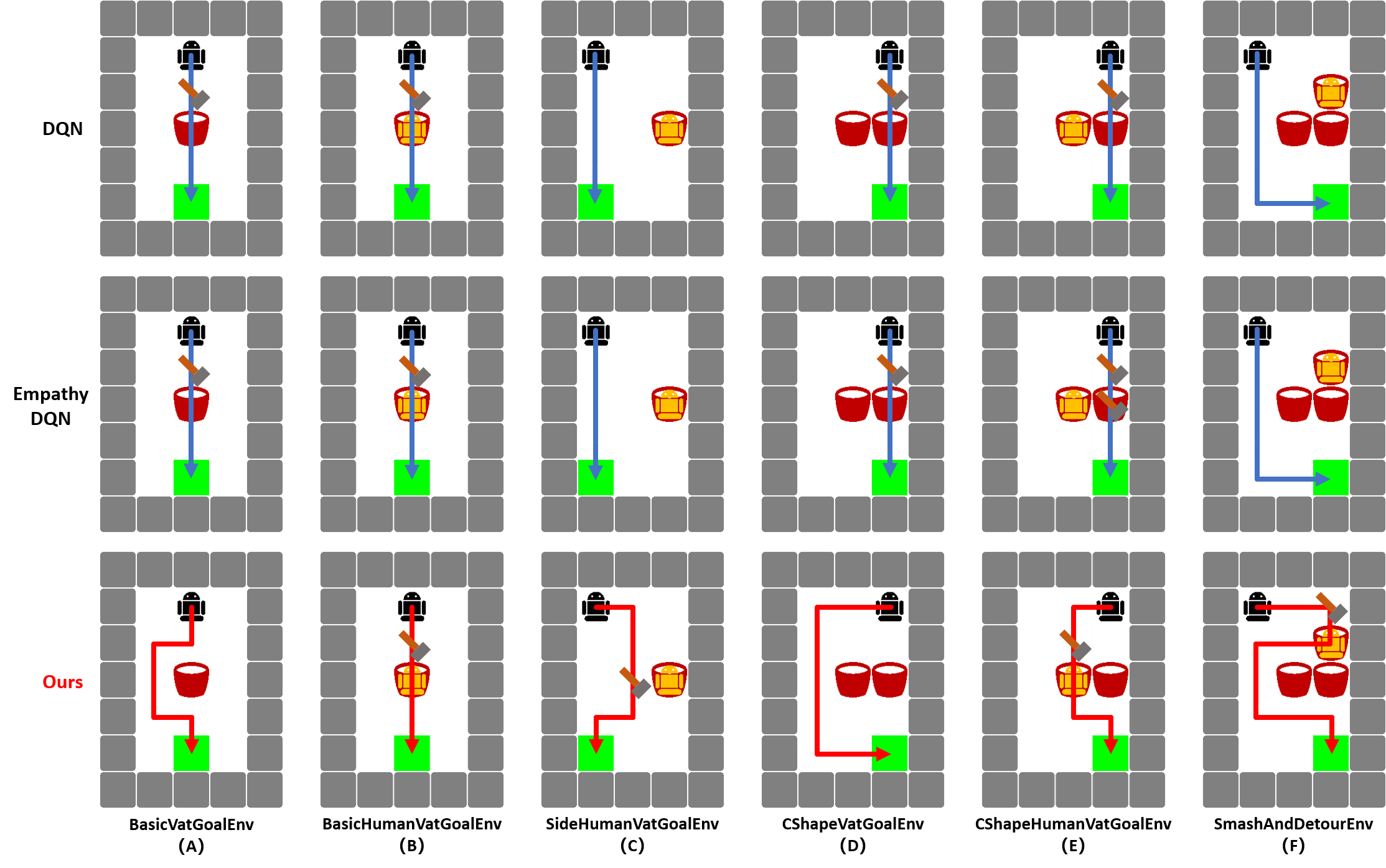}
\caption{The experimental results of different methods in various environments. We use a hammer to indicate that the agent performed a \texttt{smash} action at that position.}
\label{fig_result}
\end{figure}

Based on the fundamental smash vat environment, we also designed some variants of the task with different difficulty levels by altering the distribution of the elements, which are named as the \texttt{BasicVatGoalEnv}, \texttt{BasicHumanVatGoalEnv}, \texttt{SideHumanVatGoalEnv}, \texttt{CShapeVatGoalEnv}, \texttt{CShapeHumanVatGoalEnv} and \texttt{SmashAndDetourEnv}. 
These environments focus on different task conflicts, as shown in Table \ref{table_conf}. We tested our algorithm in these environments and the motion trajectories are shown in the last row of Fig. \ref{fig_result}. 

\begin{table}[htbp]
    \caption{Conflicts Focused in Different Environments}
    \label{table_conf}
    \centering
    \begin{tabular}{cc}
        \toprule
        \textbf{Environment} & \textbf{Conflicts} \\
        \midrule
        \texttt{BasicVatGoalEnv} & Avoid side effect vs. Agent's own task \\

        \texttt{BasicHumanVatGoalEnv} & Rescue others vs. Avoid side effects \\

        \texttt{SideHumanVatGoalEnv} & Rescue others vs. Avoid side effects vs. Agent's own task \\

        \texttt{CShapeVatGoalEnv} & Avoid side effect vs. Agent's own task \\

        \texttt{CShapeHumanVatGoalEnv} & Rescue others vs. Avoid side effects vs. Agent's own task \\

        \texttt{SmashAndDetourEnv} & Rescue others vs. Avoid side effects vs. Agent's own task \\
        \botrule
    \end{tabular}
\end{table}

From the last row of Fig. \ref{fig_result}, we can observe that the agent prioritizes rescuing people by smashing the vat, avoids the negative effects of smashing the vat as a secondary target, and ultimately finishes its own task of reaching the goal.
In the following, we use the letters A-F, as identified in Fig. \ref{fig_result}, to designate each environment.
Specifically, in the environments (A) and (D), we can discern that the agent travels a longer distance to reach the goal without smashing the vat, indicating that it has learned to avoid the environmental negative effects associated with breaking the vat. 
However, when there exist human trapped inside the vat, the agent prioritizes rescuing them above all else, as demonstrated by the outcomes in environments (B) and (E).
Furthermore, the results in environments (C) and (F) confirm that the agent is willing to take a detour to save people, indicating that an empathetic intrinsic altruistic motivation is what drives the agent to prioritize rescue efforts.
By comparing the trajectories in (B) and (C), it can be explained that the agent's action of smashing the vat in environment (C) is not for taking a shortcut to the goal, but rather its inherent empathy mechanism that determines that the priority of rescuing human is higher than avoiding environmental negative effects.

More importantly, from the results in environments (E) and (F), it can be concluded that the agent still tries to avoid smashing the vat as much as possible while rescuing people, which indicates that the overall behavior of agent is more comprehensive in terms of safety and ethics.
It is noteworthy that in (F), the agent would rather take a longer route and smash the vat to rescue trapped human from above, rather than directly reaching the left side of the trapped human to smash the vat.
This is because if agent smash the vat on the left side of the trapped human, it will also smash an additional vat that no one is trapped in (due to our setting of the \texttt{smash} action in the environment). 
So the agent would rather take a detour to avoid further impact on the environment besides rescuing Human, and this is exactly what we expected when designing the environment.

\subsubsection{Comparison with other methods}
In order to further validate the effectiveness of our method, we compared it with traditional DQN \cite{mnih2015human} as a baseline, which is trained solely on external environmental reward functions. Additionally, we compared it with Empathy DQN \cite{bussmann2019towards}, which introduced an empathy mechanism that uses agent's own strategy to speculate on the state of others.
The motion trajectories of these two methods are shown in the first and second row of Fig. \ref{fig_result}.
A concise comparison of qualitative experimental results is shown in Table \ref{table_comp_alg}.
In the table, we use a check mark ($\sqrt{}$) to indicate that the task is achieved in all six environments, a cross mark ($\times$) to indicate that the task is been achieved in all six environments, and a half check mark ($\sqrt{}\mkern-9mu{\smallsetminus}$) to indicate that the goal is achieved in some environments but not in others.

\begin{table}[htbp]
    \caption{Task Completion Status of Different Algorithms}
    \label{table_comp_alg}
    \centering
    \begin{tabular}{cccc}
        \toprule
        & Reach Goal & Avoid Side Effects & Rescue Human \\
        \midrule
        DQN\cite{mnih2015human} & $\sqrt{}$ & $\times$ &  $\times$ \\
        
        Empathy DQN\cite{bussmann2019towards} & $\sqrt{}$ & $\times$ & $\sqrt{}\mkern-9mu{\smallsetminus}$ \\
        
        Ours & $\sqrt{}$ & $\sqrt{}$ & $\sqrt{}$ \\
        \botrule
    \end{tabular}
\end{table}

Fig. \ref{fig_result} shows that, the agent trained classical DQN which solely guided by the reward function for reaching the target point, is unable to accomplish the implicit task of avoiding negative effects and rescuing trapped human. 
For agent trained by Empathy DQN, it will smash the vat to reach the target faster as it doesn't care about the irreversible impact of smashing the vat on the environment. And in some cases, it will save people along the way.
Our method can achieve all the expected goals when designing the environment.

\begin{figure}[htbp]
\centering
\includegraphics[width=1\linewidth]{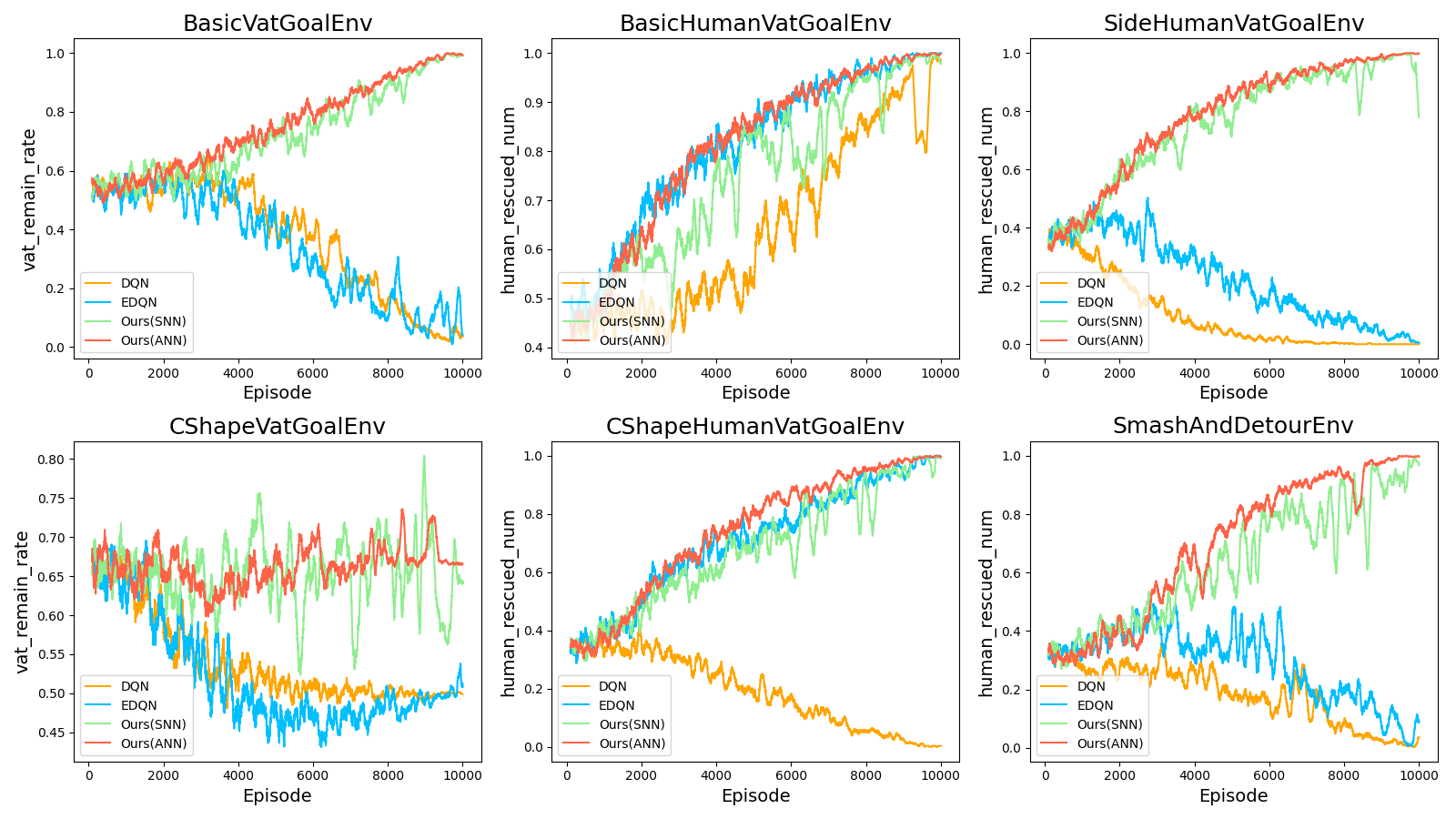}
\caption{Comparison with other methods. For each data point, we calculated the average level of the last 100 training episodes. 
We conduct 6 experiments with different random seeds and take the average values.}
\label{fig_comp}
\end{figure}

To conduct a further quantitative comparison, we plotted the curves showing the changes in the agent's impact on the environment (the average remaining vat rate) and the rescue situation (the average number of human rescued) during the training process, as shown in Fig. \ref{fig_comp}.
In the \texttt{BasicVatGoalEnv} and \texttt{CShapeVatGoalEnv}, where there are no trapped human, our focus is solely on the average vat remaining rate at the end of the training process. In other environments, we are more concerned with the average number of people rescued per episode over the recent 100 episodes.

The curves in Fig. \ref{fig_comp} show that neither classic DQN nor Empathy can effectively avoid negative effects and altruistic rescue, while our method demonstrates superiority across different environments.
The agent trained by classic DQN rescued human in \texttt{BasicHumanVatGoalEnv}, but by comparing the results with other environments, it can be deduced that this was a result of smashing the vat to take a shortcut and reach the target point more quickly, rather than being driven by empathy.
The agent trained by Empathy DQN seems to perform quite well in the \texttt{BasicHumanVatGoalEnv} and \texttt{CShapeHumanVatGoalEnv}, which is because in these two environments the vat where human is trapped in is close to the shortest path from the starting position to the target.
When the trapped human needs the agent to take a detour to rescue, it performs poorly, as shown in the experimental result in \texttt{SideHumanVatGoalEnv} and \texttt{SmashAndDetourEnv}, which indicates that the empathetic capability of Empathy DQN is not strong enough.

\subsubsection{Ablation Experiment}

We also conducted ablation experiments to verify the respective effects of the penalty term for negative effects $R_{nse}(s_t,a_t)$ defined in Eq. \ref{eq_rnse} and the incentive term for empathizing with others $R_{emp}(s_t,a_t)$ defined in Eq. \ref{eq_remp} in our proposed method. 
We compare the average vat remaining rate and human rescued rate in the last 100 training episode where the algorithm converges and the results tend to be stable, the relevant results are shown in Table \ref{table_ablation}.

\begin{table}[htbp]
    \caption{Comparison of Task Completion Status of Different Combination of Rewards }
    \label{table_ablation}
    \centering
    \begin{tabular}{cccccc}
        \toprule
         & & $R_{env}$ & $R_{env}+R_{nse}$ & $R_{env}+R_{emp}$ & $R_{total}$ \\
        \midrule
        \multirow{2}{*}{\texttt{BasicVatGoalEnv}} &
        vat remain rate & 0.038 & 0.995 & 0.007 & 0.992 \\ 
        & human rescue rate & - & - & - & - \\
        \midrule
        \multirow{2}{*}{\texttt{BasicHumanVatGoalEnv}} &
        vat remain rate & 0.013 & 0.997 & 0.000 & 0.002 \\ 
        & human rescue rate & 0.987 & 0.003 & 1.000 & 0.998 \\
        \midrule
        \multirow{2}{*}{\texttt{SideHumanVatGoalEnv}} &
        vat remain rate & 1.000 & 1.000 & 0.002 & 0.002 \\ 
        & human rescue rate & 0.000 & 0.000 & 0.998 & 0.998 \\
        \midrule
        \multirow{2}{*}{\texttt{CShapeVatGoalEnv}} &
        vat remain rate & 0.499 & 0.666 & 0.497 &  0.666 \\ 
        & human rescue rate & - & - & - & - \\
        \midrule
        \multirow{2}{*}{\texttt{CShapeHumanVatGoalEnv}} &
        vat remain rate & 0.502 & 0.733 & 0.498 & 0.502 \\ 
        & human rescue rate & 0.003 & 0.002 & 0.998 & 0.995 \\
        \midrule
        \multirow{2}{*}{\texttt{SmashAndDetourEnv}} &
        vat remain rate & 0.973 & 1.000 & 0.561 & 0.519 \\ 
        & human rescue rate & 0.035 & 0.000 & 1.000 & 0.997 \\
        \botrule
    \end{tabular}
\end{table}

The results of the ablation experiment demonstrate the effectiveness and necessity of the proposed $R_{nse}(s_t,a_t)$ and $R_{emp}(s_t,a_t)$ in our method. 
When there is only environmental reward $R_{env}$, the method degenerates into the classic DQN algorithm, which is solely oriented towards the goal regardless of environmental negative effects and altruistic rescue. 
Whenever the vat is on the shortest path from the starting position to the goal, the agent trained by DQN will directly smash the vat and head towards the goal, regardless of whether there are people trapped inside the vats.
When only considering the integration of negative effect penalties $R_{nse}$ and environmental rewards $R_{env}$, the vat remaining rate high, but the agent hardly rescues human. Therefore, it is difficult for it to solve conflict decision-making environments that require rescuing people trapped in vats at the cost of damaging the environment.
When considering only the the empathy incentive term $R_{nse}$, the agent is unable to handle environments with no human presence.

When environmental reward feedback $R_{env}$, negative effect punishment $R_{nse}$ and empathy incentive term $R_{emp}$ are combined, the resulting trained agent can effectively handle all the aforementioned scenarios.
In environments where no one is trapped, the agent will avoid smashing the vat; When there is a conflict between smashing the vat and rescuing trapped human, the agent will prioritize smashing the vat to save trapped human; And when there exist multiple vats in the environment, but only some contain trapped human, the agent will only smash the vats with trapped people, then bypass the other vats and head towards the goal.

An interesting and noteworthy observation is that when only components $R_{env}$ and $R_{emp}$ are integrated, the agent achieves nearly the same performance as the full integration of $R_{env}$, $R_{nse}$, and  $R_{emp}$ in environments where exists trapped human.
This possibly suggests that, apart from saving lives, refraining from unnecessarily breaking vats also aligns with the interests of others.
This may imply that refraining from unnecessarily breaking vats also aligns with the interests of others.

In general, $R_{emp}(s_t,a_t)$ can motivate agents to prioritize rescuing people in the presence of trapped humans, and $R_{nse}(s_t,a_t)$ ensures agents to avoid negative effects when there are no people in the environment.

\subsubsection{Hyperparameter Analysis}

\begin{figure}[htbp]
\centering
\includegraphics[width=1\linewidth]{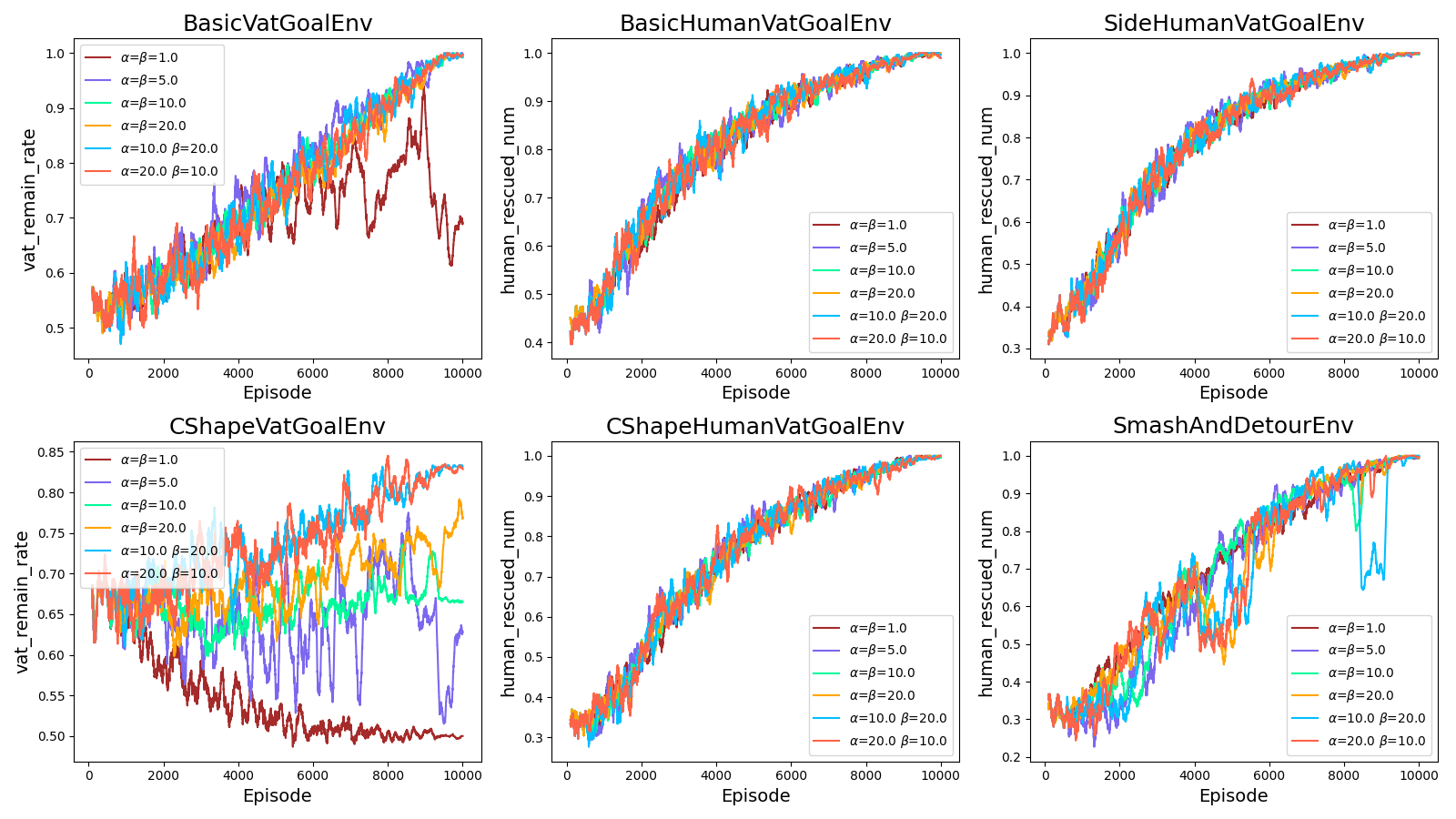}
\caption{Hyperparameter experiment result. The data processing method is similar to Fig. \ref{fig_comp}}
\label{fig_weight}
\end{figure}

We also tested the impact of hyperparameters $\alpha$ and $\beta$ proposed in Eq. \ref{eq_rtotal} on our algorithm, which are used to control the agent‘s tendency to avoid negative effects and empathetic altruism.
How the relative weights of environmental rewards, negative effect penalties and empathy altruism incentive affect the behavior of the final trained agent is a question worth exploring.
Thus, we conducted tests by selecting several typical values within the range of [1, 20] while keeping $\alpha=\beta$, and also tested two scenarios where $\alpha$ and $\beta$ are not equal.
The comparison result is plotted in Fig. \ref{fig_weight} using a method similar to plotting Fig. \ref{fig_comp}.

In most environments, the result curves under different hyperparameter settings almost coincide, which indicate that the value of the hyperparameter within a reasonable range does not significantly affect the experimental results, the agent is capable of prioritizing rescue based on an empathetic mechanism and avoiding smashing the vat when no one is trapped inside. This shows the robustness of our proposed method.
An excessively small hyperparameter value ($\alpha=\beta=1.0$) may result in insufficient punishment for the vat-smashing behavior and allowing the agent to smash the vat to reach the goal quicker, which is reflected in the experimental result of \texttt{BasicVatGoalEnv} and \texttt{CShapeVatGoalEnv}, where exists no trapped human.

\subsubsection{The Compatibility on SNN and DNN}
In order to test the compatibility of our proposed random imagination based method with different network models, inspired by some computational empathy models based on SNN that incorporate brain-inspired mechanisms, we have also considered integrating our approach with spiking neural networks \cite{maass1997networks} and testing its efficacy. 
We adopt a direct spike encoding strategy. By replacing the ReLU neurons \cite{nair2010rectified} in our network with Leaky Integrate-and-Fire (LIF) neurons \cite{dayan2005theoretical}, we substituted the original DNN with an SNN without changing the network architecture\cite{zeng2023braincog}. We used the surrogate gradient backpropagation algorithm \cite{shen2022backpropagation} to optimize the network during the training process. 
More detailed information about SNN can be found in Appendix \ref{app_snn}.
The corresponding results have been depicted in the green lines of Fig. \ref{fig_comp}.

The experimental results show that our performs well when integrated with SNN.
It is a natural outcome considering that our proposed method is essentially model independent.
This indicates that the intrinsic motivation mechanism we proposed can be easily integrated with other existing deep reinforcement learning algorithms regardless of their specific network architectures, demonstrating its broad applicability and robustness.

\section{Discussion}

Drawing inspiration from the ancient Chinese story of \textit{Sima Guang Smashes the Vat} and human cognitive ability, this paper proposes a unified computational framework of self-imagination integrated with ToM, empowering agents to autonomously align with human values on altruism.
This framework enables agents to predict the potential impacts of their actions, particularly by using perspective-taking to forecast the effects of their decisions on others' interests, before making decisions, thus generating intrinsic motivations leading to safe, moral, and altruistic decision-making.
We design an experimental scenario similar to \textit{Sima Guang Smashes the Vat} and its variants with different complexities, where exist conflicts among agents' own task, rescuing others and avoiding negative effects.
Experimental results show that the agent trained by our proposed method is able to balance the above contradictions, prioritize rescuing individuals while minimizing environmental negative impacts and completing their own tasks. 
Further experiments demonstrate the effectiveness of the proposed framework as well as its good robustness under different hyperparameter configurations and compatibility with different networks.

Our proposed method differs from some existing related methods. Below is a brief description of the main differences between our method and existing ones, along with the advantages our method offers:
\begin{enumerate}
    \item 
    Compared with existing pure RL methods, such as DQN \cite{mnih2015human}, our method is capable of considering the impact of agent actions on the environment and others, thereby generating internal incentives to avoid negative effects and empathize with altruism without explicit specified reward function. 
    This can prompt agents to make more ethical and safe decisions without external reward function guidance.

    \item 
    In comparison to existing methods that only account for the negative impact of an agent on the environment, such as original AUP \cite{turner2020conservative} and FTR \cite{krakovna2020avoiding}, our approach additionally incorporates empathy towards others. 
    This overcomes the limitation of agent being overly conservative that only avoid negative effects, enabling the agent to make proactive decisions to aid others, even at the cost of causing irreversible damage to the environment.

    \item 
    Compared to existing methods that only consider empathy towards others, such as Empathy DQN \cite{bussmann2019towards}, our approach avoids the negative effects of environmental destruction without the need for an explicitly defined reward function, especially in those environments where there are no empathizable subjects, thereby exhibiting greater universality and generalizability.

    \item 
    Compared with the work of Alamdari et al. \cite{alizadeh2022considerate}, which extended existing methods to avoid negative effects (FTR\cite{turner2020avoiding}) to enable agents to empathize with others and thus avoid harming others' interests, our method estimate others' status based on self-experience and does not require obtaining rewards from others, thus providing wider applicability.
    
\end{enumerate}

We aspire to ultimately enable AI to comprehend human morality, so that it can better benefit human society.
The significance of this work lies more in a preliminary exploration of agents autonomous alignment with human altruistic values, laying the foundation for the subsequent realization of moral and ethical AI.
Nevertheless, the environment we have designed remains insufficiently complex. Given the complexity of altruistic motivations in humans within intricate social environments \cite{wu2024motive, jin2024observing}, future research needs to design more sophisticated experimental settings.
Besides, the proposed intrinsic incentive mechanism is not predicated on enabling agents to comprehend human morality. 
Looking ahead, we intend to further explore more intricate and conflict-ridden decision-making environments, and contemplate utilizing powerful tools like large language models, which possess significant representational and comprehension capabilities, to endeavor to enable agents to exhibit safer, more ethical, and altruistic decision-making behaviors, all while aligning with human moral values.

\section{Methods}
When an agent performs a task without an explicit external reward, it shows indifference to the negative impact of its behavior on the environment or on the interests of other agents.
To align with human altruistic values, and achieve safe and altruistic behavior that generalizes across different situations, especially when there is a conflict between the agent's task, environmental negative effects, and the interests of other agents, intrinsic incentive generation mechanisms are essential. 
Intrinsic safety and moral altruistic behavior arise from imagining the potential impact of actions on the environment and others based on the agent's own experience.  
Based on this, the model proposed in this paper consists of three main components: the agent's imaginary space  updated based on the its self-experiences, intrinsic motivation to avoid negative effects and empathy towards others by perspective taking, along with the interaction and coordination between the self-imagination module and decision-making network. 
The overall framework of our proposed model can be seen as Fig.~\ref{fig_alg}.
The relevant implementation code can be found at \url{https://github.com/BrainCog-X/Brain-Cog/tree/main/examples/Social_Cognition/SmashVat}.

\subsection{Self-imagination Module}
In everyday life, humans frequently anticipate the consequences of their decisions before acting. 
Considering a simple example, when a mother asks her child to sweep the floor, the child may consciously avoid areas such as a table with water bottles even without explicit instructions. 
This behavior arises from the child’s ability to mentally simulate potential outcomes, such as accidentally knocking over the bottle, which could result in upsetting the mother or necessitate additional time and effort to clean up spilled water and broken bottle fragments. 
Although these imagined scenarios do not actually occur, they significantly influence the child’s decision-making process. 
This observation inspires the idea of enabling agents to make rational decisions by endowing them with the capacity to imagine possible outcomes based on their prior experiences.

% In the example of a child sweeping the floor mentioned above, the imagined outcome of the child relies on their prior knowledge of the environment, such as the fragility of vases, and this imagination is highly correlated with the environment. 
% We hope that our algorithm can have strong generality, so when implementing the imaginative ability given to agents, we expect that it does not require prior knowledge and is decoupled from the environment. 
% Inspired by AUP \cite{turner2020conservative}, we generate a group of imaginary state-action space, which are independent virtual environments that exactly copied from the real environment but with different randomly generated reward functions, to achieve the agent's imagination ability.
Self-imagination can be implemented through various specific approaches.
Inspired by the work of AUP \cite{turner2020conservative}, we adopt a method based on random reward functions. 
This implementation offers several advantages: 
random number generation offers simplicity and efficiency at the algorithmic level compared to complex reward generation mechanisms while enabling coverage of diverse scenarios;
and it eliminates the need for prior knowledge about the environment, and thus decouples from specific environmental tasks, providing robust generalization capabilities.

Formally, the interaction between agent and the real environment can be modeled as a MDP $\langle \mathcal{S}, \mathcal{A}, T, R, \gamma \rangle$ with state space $\mathcal{S}$, action space $\mathcal{A}$, transition function $T:\mathcal{S} \times \mathcal{A} \rightarrow \Delta(\mathcal{S})$, reward function $R:\mathcal{S} \times \mathcal{A} \rightarrow \mathbb{R}$, and discount factor $\gamma \in [0,1)$. 
The environment of the imaginative space is based on the real environment, except that it employs randomized rewards that are independent of the environment, which means activities in different independent imaginative spaces can be modeled as $\langle \mathcal{S}, \mathcal{A}, T, R_i, \gamma \rangle \quad i=1,2,\dots,N$, where $R_i$ randomly generated reward function conforming to the uniform distribution of $[0,1)$ , and $N$ is specified number of imaginary environments. 

Since the activities in the imaginary spaces are also MDPs, we maintain a learnable Q-value function $Q_i$ in each imaginary space to estimate the values of various states in imagined situations.
Notably, the transition functions \( T \) in these imaginary spaces are identical to those in the real environment, thus we do not actually establish multiple separate imagination spaces or learn each $Q_i$ through state transitions $T(s_t, a_t, s_{t+1})$ generated from the interaction between agent and these spaces. 
In fact, each $Q_i$ is learned through the agent's direct interactions with the real environment, illustrating that the agent's imagination is based on real-world experience.
For each transaction $T(s_t, a_t, s_{t+1})$, $Q_i$ is updated using Eq.\ref{eq_updateq}
\begin{equation}
    \label{eq_updateq}
    Q_i(s_t, a_t)\leftarrow 
    \max \left[
        Q_i(s_t, a_t), \
        R_i(s_t, a_t) + \gamma \max \limits_{a'}{Q_i(s_{t+1}, a')}
    \right]
\end{equation}
which differs from the original AUP method that utilizes Q-learning to update these Q-value functions.
$Q_i$ can be seen as the quantification of the consequences of actions in the imaginary space, thus it is what we actually use in the following calculation.

\subsection{Avoid Negative Side Effects}
With the estimated values of various states in imagined situations $Q_i$, we can utilize them to enable the agent anticipate the potential consequences of its decision-making actions before execution, thus avoiding negative environmental effects.
An intuitive idea is that the agent should imagine the possible consequences of taking a certain action $a$ at the current state $s$ by examining the specific Q-value $Q_i(s,a)$ to determine whether the consequences are good or bad. 
However, the concept of good or bad is relative, and only becomes meaningful when there exists a reference for comparison. 
Hence, we consider introducing a baseline state to serve as a comparative standard.

\begin{figure}[htbp]
\centering
\includegraphics[width=0.6\linewidth]{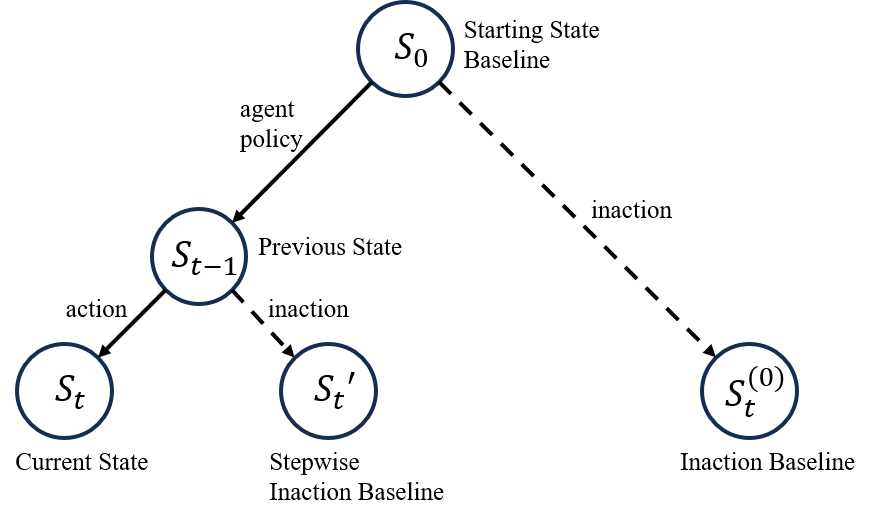}
\caption{The relationship between different baselines.}
\label{fig_baseline}
\end{figure}

Fig. \ref{fig_baseline} shows different choices of baselines. 
Like Krakovna \cite{krakovna2018penalizing} and Turner \cite{turner2020avoiding} et al., we choose the stepwise inaction state $S_t'$ as the baseline state. 
One natural choice of baseline is the starting state $S_0$, but this might cause a penalty on change of the environment that is not caused by the agent's action. 
To avoid this, the inaction baseline $S_t^{(0)}$ seems to be a more reasonable choice of baseline called, which is referred to the state that the environment would currently be in if the agent have never acted. 
But inaction baseline may cause other problems.
Using this baseline state may lead to the continuous accumulation of penalties or incentives resulting from certain behaviors, ultimately yielding incorrect outcomes.
Therefore, we use the stepwise inaction baseline, which can avoids penalizing the effects of a single action multiple times and ensures that not acting incurs zero penalty.

Here we use empty set symbol $\emptyset$ to indicate that the agent's inaction, then the imaginary change of the environment caused by action $a$ under state $s$ can be expressed as $Q_i(s,a)-Q_i(s,\emptyset)$. Since we just want to punish those actions that cause negative effects on the environment, so we define the negative side effect penalty term $R_{nse}(s,a)$ as follows:
\begin{equation}
\label{eq_rnse}
    R_{nse}(s,a) := 
    \frac{1}{N} \sum\limits_{i=1}^N
    {\left| \min \left(0,\ Q_i(s,a)-Q_i(s,\emptyset)\right)\right|}
\end{equation}
which is an average of all negative changes caused by the action of different Q-value functions.

\subsection{Self-experience based ToM}

The key to achieving ToM lies in considering problems in the shoes of others.
In the context of evaluating the potential impacts of decision-making actions on the environment, agents can extend their considerations further by accounting for how environmental changes may affect others.
However, directly inputting the observed current state of others into the agent's own strategy network to estimate the impact of actions on them \cite{bussmann2019towards} implicitly assumes that the agent and others share similar tasks.
Consequently, this approach lacks generalization in environments where the tasks of the agent and others are inconsistent.
Obviously, if we can directly obtain rewards and estimated value of the current state from others and incorporate them into the agent's decision-making\cite{alizadeh2022considerate}, it would be beneficial for the agent to make altruistic decisions. 
However, in the real environment, it is difficult for us to obtain task rewards and value estimates from others for the state.
And using inverse reinforcement learning to estimate others' tasks and rewards \cite{senadeera2022sympathy} is too complex and computationally time-consuming.

% 可能还要再改解释
% Essentially, the self-imagination method works by comparing the agent’s state reachability (i.e., the number of states that can be reached) before and after executing an action. Therefore, it is well-suited for taking the perspective of others, assessing their state reachability, and learning how one’s own actions affect them, thereby fostering altruistic motivation. We employ the Q-value functions $Q_i$ learned based on randomly generated reward in the imaginary virtual environments mentioned above to to implement empathy.

Since $Q_i$ are learned based on randomly generated reward, they are decoupled from the real task reward of the environment.
Thus, we  use $Q_i$ to estimate the value of others' states to achieve empathy, which can avoid errors caused by inconsistencies between the agent and others' real tasks.
Although the agent and others may have different tasks, they share the same environment and the expected outcome of interacting with the environment is similar. 
Therefore, it is reasonable to directly use the same $Q_i$ to estimate the value of others' state $Q_i(s^{others}, a)$, which reflects the essence of empathy.
By adopting this approach, we have unified the avoidance of negative effects and empathy altruism into the same computational framework of self-imagination, thereby avoiding the extra computing cost of using different methods to calculate empathy altruism incentive term.

As mentioned before, the agent should consider the effects of environmental changes caused by action $a$ of the agent on others while making decisions, the changes can be represented by $Q_i(s^{others},a)-Q_i(s^{others},\emptyset)$.
Thus, we define the inherent empathy incentive term $R_{nse}(s,a)$ as follows:
\begin{equation}
\label{eq_remp}
    R_{emp}(s,a) := \frac{1}{N} \sum\limits_{i=1}^N{(Q_i(s^{others},a)-Q_i(s^{others},\emptyset))} 
\end{equation}
which is an average of all changes of different Q-value functions to encourage agents to perform actions that benefit others while suppressing actions that are detrimental to others.

\subsection{Integration of Self-imagination Module and Decision-making Network}

The negative effects and empathy-related rewards generated within the imagined space serve directly as intrinsic rewards that influence the agent’s decision-making. In other words, when making decisions in the real environment, the agent comprehensively considers both the actual environmental feedback $R_{env}$ and the intrinsic rewards predicted in its imagined space. Using the total reward function $R_{total}$, the DQN network is then optimized to adjust the decision-making strategy accordingly. Based on the definition of $R_{nse}$ given by Eq.\ref{eq_rnse} and $R_{emp}$ given by Eq. \ref{eq_remp}, here we propose the complete reward function $R_{total}$ as follows:

\begin{equation}
\label{eq_rtotal}
    R_{total}(s,a) := \frac{R_{env}(s,a) - \alpha R_{nse}(s,a) + \beta R_{emp}(s,a) }{(\alpha+\beta)/2} 
\end{equation}
where $R_{env}(s,a)$ refers to the original environmental reward function, $\alpha$ and $\beta$ denote weight hyperparameters used to control the tendency of agents.

The imaginary space and the real environment interact in real time, forming a dynamic and positive loop. 
The state transitions $(s_t, a_t, s_{t+1})$ generated from the interaction between the agent and real environment are consistently used to update the imagined space, while the intrinsic motivation derived from this imagined space guides the decision-making module in executing safe and moral behavior. 
This positive real-time interaction facilitates synchronized online learning via a shared self-experience buffer.

The complete algorithm process is shown in pseudocode Algorithm \ref{alg1}.

\begin{algorithm}[ht]
    \caption{Avoid negative side effect with empathy}
    \label{alg1}
    \begin{algorithmic}[1]
        \State Init policy net $Q_{policy}$ with weight $\theta$;
        \State Init target net $Q_{target}$ with weight $\theta'=\theta$;
        \State Init self-experience replay buffer;
        \State Generate the random functions $R_1, R_2, \dots , R_N$ of $N$ 
        \Statex different imaginary environments;
        \State Init state value estimation $Q_i\quad i=1,2,\dots,N$ for each imaginary environment;
        \For{$episode=1, 2,\dots, N_{num\_episode}$}
            \State Reset environment and get initial states $s_1$ and $s_1^{others}$;
            \For{$t=1,2,\dots, T_{maxstep}$}
                \State Select a random action $a_t$ with probability $\epsilon$ 
                \Statex \qquad \quad or select $a_t = \arg\max\limits_a Q_{policy}(s_t,a)$;
                \State Apply action $a_t$ to the environment and 
                \Statex \qquad \quad get reward $r_t$ and next state $s_{t+1}$ $s_{t+1}^{others}$;
                \State Update each $Q_i$ with transaction $(s_t,a_t, s_{t+1})$ using Eq. \ref{eq_updateq};
                \State Calculate the side effect penalty term $R_{nse}(s,a)$ using Eq. \ref{eq_rnse};
                \State Calculate the empathy incentive term $R_{emp}(s,a)$ using Eq. \ref{eq_remp};
                \State Calculate the total reward $R_{total}(s,a)$ using Eq. \ref{eq_rtotal};
                \State Store transaction $(s_t,a_t,r_t^{total}, s_{t+1})$ in replay buffer;
                \State Sample a batch of transaction from replay buffer and optimize $Q_{policy}$;
                \State Every several steps set $\theta' = \theta$;
            \EndFor
        \EndFor
    \end{algorithmic}
\end{algorithm}

\backmatter

\bmhead{Acknowledgements}
This work was supported by the Strategic Priority Research Program of the Chinese Academy of Sciences (Grant No. XDB1010302), the Beijing Major Science and Technology Project(Grant No. Z241100001324005), the National Natural Science Foundation of China (Grant No. 62106261 and No. 32441109) and the funding from Institute of Automation, Chinese Academy of Sciences(Grant No. E411230101).

\begin{appendices}

\section{Experiment Details}
\label{app_exp_detail}
Here we supplement additional details on the experimental setup that were not thoroughly described in the main text.

\subsection{The Observation Space of the Smash Vat Environment}
In the smash vat environment, the agent possesses global observation capabilities, which means the agent can observe every object in the environment, the location of every other agent (if there exist), as well as its own position.

Specifically, the observation space of the agent is a $3\times7\times5$ image-like array, or tensor.
The first channel of the array represents the distribution of various elements in the environment, using 0 to denote an empty grid and integers from 1 to 3 to represent other elements besides \texttt{human}.
The second channel uses one-hot encoding to represent the agent's current position, where the value at the position corresponding to the agent's location is 1, and all other positions have a value of 0.
The third channel uses one-hot encoding to represent the location of other agents.

From another perspective, each grid in the grid world corresponds to a triplet $(a, b, c)$, where $a$ represents the attributes of the grid, $b$ indicates the presence of an agent, and $c$ denotes the presence of an other agent.

\subsection{Network Architecture}
The Architecture of the network used in our method is shown as Table \ref{table_net}.

\begin{table}[htbp]
    \caption{Architecture of the Network}
    \label{table_net}
    \centering
    \begin{tabular}{cccccc}
        \toprule
        \textbf{Layer} & \textbf{Input Size} & \textbf{Kernel} & \textbf{Stride} & \textbf{Padding} & \textbf{Output Size} \\
        \midrule
        Conv 1 & $3 \times 7 \times 5$ & $3\times3$ & 1 & 1 &  $16 \times 7 \times 5$ \\
        
        Conv 2 & $16 \times 7 \times 5$ & $3\times3$ & 1 & 0 &  $32 \times 5 \times 3$ \\
        
        Conv 3 & $32 \times 5 \times 3$ & $3\times3$ & 1 & 0 &  $64 \times 3 \times 1$ \\
        \midrule
        AvgPool & $64 \times 3 \times 1$ & - & - & - &  $64 \times 1 \times 1$ \\
        
        Flatten & $64 \times 1 \times 1$ & - & - & - &  64 \\
        \midrule
        Linear 1 & 64 & - & - & - &  128 \\
        
        Linear 2 & 128 & - & - & - &  6 \\
        \botrule
    \end{tabular}
\end{table}

The DNN and SNN we used in the experiment share the same network structure shown in Table \ref{table_net}. The only difference is that DNNs employ ReLU neurons, while SNNs utilize LIF neurons.

\subsection{Training Hyperparameter Settings}
The relevant hyperparameters used in training are shown in Table \ref{table_hyper}.

\begin{table}[htbp]
    \caption{Training Hyperparameters}
    \label{table_hyper}
    \centering
    \begin{tabular}{cc}
        \toprule
        \textbf{Name} & \textbf{Value} \\
        \midrule
        replay buffer size & 100000\\
        batch size & 100 \\
        target net update interval &  1000 \\
        learning rate & 0.0001 \\
        training episodes & 10000 \\
        $\gamma$ & 0.99 \\
        number of imaginary spaces & 30 \\
        \botrule
    \end{tabular}
\end{table}

When training the policy network, we employed an $\epsilon$-greedy strategy for action selection. The specific decay strategy for $\epsilon$ is as follows: 
In the first 500 episodes of training, we maintain an $\epsilon$ at a value of 1.00 to ensure the agent fully explores the environment. Then, $\epsilon$ linearly decays to 0.01 and is finally maintained at 0.01 for the last 500 episodes of training.

\section{Spiking Neural Network}
\label{app_snn}
Here we provide a brief introduction to SNN.

Spiking Neural Network, as the third generation of neural networks \cite{maass1997networks}, is a more biologically plausible model of neural networks. Compared to traditional DNN, SNN emphasizes the use of spike sequences with precise firing times as the basic carriers of information.

\subsection{LIF Neuron}
The LIF neuron \cite{dayan2005theoretical} is a simplified model that describes the generation and propagation mechanism of neuronal action potentials. 
It abstracts the cell membrane as an equivalent circuit containing a capacitor, resistor and power source, where the capacitor reflects the capacitance of the cell membrane, the resistor reflects the permeability of the leak channels, and the power source reflects the influence of external input currents and the resting potential.

The differential equation describing the LIF neuron is as Eq. \ref{eq_lif}:
\begin{equation}
\label{eq_lif}
    \tau \frac{du}{dt} = -[u(t)-u_{rest}] + R I(t)
\end{equation}
where $u(t)$ denotes membrane potential, $u_{rest}$ denotes the the resting potential, $I(t)$ denotes the input currents, $\tau = RC$ denotes the time constant, and $R$ and $C$ denote the membrane resistance and capacitance, respectively.

\subsection{Direct Spike Encoding Strategy}
Information is transmitted between neurons in the form of spike sequences, thus requiring a specific encoding scheme to encode the input as a series of spike sequences.

Direct spike encoding strategy duplicates the input multiple times, with each copy corresponding to a time step, and then inputs them into the network sequentially. 
Direct encoding can be viewed as applying a constant current stimulus to the neurons in the first layer \cite{rueckauer2017conversion}. 
These neurons will generate corresponding pulse sequences based on their own dynamic characteristics and synaptic weights, and transmit them to subsequent layers. 
In this case, the first layer acts as a learnable encoder that can adjust its parameters based on feedback from network training, thereby achieving optimal encoding of the analog input signal.

\subsection{Surrogate Gradient Backpropagation}
Because of the nondifferentiable nature of the spiking function, the gradient of a smoother function called the surrogate gradient is used as an alternative to the real gradient, enabling the back propagation algorithm to be successfully applied to the training of SNNs \cite{shen2022backpropagation}.
The surrogate gradient function used in our experiment to replace the spiking function is defined in Eq. \ref{eq_sur_func} .
\begin{equation}
\label{eq_sur_func}
    g(x) =
        \begin{cases}
        0, & x < -\frac{1}{\alpha} \\
        -\frac{1}{2}\alpha^2|x|x + \alpha x + \frac{1}{2}, & |x| \leq \frac{1}{\alpha}  \\
        1, & x > \frac{1}{\alpha} \\
        \end{cases}
\end{equation}

\end{appendices}

\bibliography{ref}


\begin{thebibliography}{42}
\ifx \bisbn   \undefined \def \bisbn  #1{ISBN #1}\fi
\ifx \binits  \undefined \def \binits#1{#1}\fi
\ifx \bauthor  \undefined \def \bauthor#1{#1}\fi
\ifx \batitle  \undefined \def \batitle#1{#1}\fi
\ifx \bjtitle  \undefined \def \bjtitle#1{#1}\fi
\ifx \bvolume  \undefined \def \bvolume#1{\textbf{#1}}\fi
\ifx \byear  \undefined \def \byear#1{#1}\fi
\ifx \bissue  \undefined \def \bissue#1{#1}\fi
\ifx \bfpage  \undefined \def \bfpage#1{#1}\fi
\ifx \blpage  \undefined \def \blpage #1{#1}\fi
\ifx \burl  \undefined \def \burl#1{\textsf{#1}}\fi
\ifx \doiurl  \undefined \def \doiurl#1{\url{https://doi.org/#1}}\fi
\ifx \betal  \undefined \def \betal{\textit{et al.}}\fi
\ifx \binstitute  \undefined \def \binstitute#1{#1}\fi
\ifx \binstitutionaled  \undefined \def \binstitutionaled#1{#1}\fi
\ifx \bctitle  \undefined \def \bctitle#1{#1}\fi
\ifx \beditor  \undefined \def \beditor#1{#1}\fi
\ifx \bpublisher  \undefined \def \bpublisher#1{#1}\fi
\ifx \bbtitle  \undefined \def \bbtitle#1{#1}\fi
\ifx \bedition  \undefined \def \bedition#1{#1}\fi
\ifx \bseriesno  \undefined \def \bseriesno#1{#1}\fi
\ifx \blocation  \undefined \def \blocation#1{#1}\fi
\ifx \bsertitle  \undefined \def \bsertitle#1{#1}\fi
\ifx \bsnm \undefined \def \bsnm#1{#1}\fi
\ifx \bsuffix \undefined \def \bsuffix#1{#1}\fi
\ifx \bparticle \undefined \def \bparticle#1{#1}\fi
\ifx \barticle \undefined \def \barticle#1{#1}\fi
\bibcommenthead
\ifx \bconfdate \undefined \def \bconfdate #1{#1}\fi
\ifx \botherref \undefined \def \botherref #1{#1}\fi
\ifx \url \undefined \def \url#1{\textsf{#1}}\fi
\ifx \bchapter \undefined \def \bchapter#1{#1}\fi
\ifx \bbook \undefined \def \bbook#1{#1}\fi
\ifx \bcomment \undefined \def \bcomment#1{#1}\fi
\ifx \oauthor \undefined \def \oauthor#1{#1}\fi
\ifx \citeauthoryear \undefined \def \citeauthoryear#1{#1}\fi
\ifx \endbibitem  \undefined \def \endbibitem {}\fi
\ifx \bconflocation  \undefined \def \bconflocation#1{#1}\fi
\ifx \arxivurl  \undefined \def \arxivurl#1{\textsf{#1}}\fi
\csname PreBibitemsHook\endcsname

%%% 1
\bibitem[\protect\citeauthoryear{Amodei et~al.}{2016}]{amodei2016concrete}
\begin{botherref}
\oauthor{\bsnm{Amodei}, \binits{D.}},
\oauthor{\bsnm{Olah}, \binits{C.}},
\oauthor{\bsnm{Steinhardt}, \binits{J.}},
\oauthor{\bsnm{Christiano}, \binits{P.}},
\oauthor{\bsnm{Schulman}, \binits{J.}},
\oauthor{\bsnm{Man{\'e}}, \binits{D.}}:
Concrete problems in ai safety.
arXiv preprint arXiv:1606.06565
(2016)
\end{botherref}
\endbibitem

%%% 2
\bibitem[\protect\citeauthoryear{Leike et~al.}{2017}]{leike2017ai}
\begin{botherref}
\oauthor{\bsnm{Leike}, \binits{J.}},
\oauthor{\bsnm{Martic}, \binits{M.}},
\oauthor{\bsnm{Krakovna}, \binits{V.}},
\oauthor{\bsnm{Ortega}, \binits{P.A.}},
\oauthor{\bsnm{Everitt}, \binits{T.}},
\oauthor{\bsnm{Lefrancq}, \binits{A.}},
\oauthor{\bsnm{Orseau}, \binits{L.}},
\oauthor{\bsnm{Legg}, \binits{S.}}:
Ai safety gridworlds.
arXiv preprint arXiv:1711.09883
(2017)
\end{botherref}
\endbibitem

%%% 3
\bibitem[\protect\citeauthoryear{Park et~al.}{2024}]{park2024ai}
\begin{botherref}
\oauthor{\bsnm{Park}, \binits{P.S.}},
\oauthor{\bsnm{Goldstein}, \binits{S.}},
\oauthor{\bsnm{O’Gara}, \binits{A.}},
\oauthor{\bsnm{Chen}, \binits{M.}},
\oauthor{\bsnm{Hendrycks}, \binits{D.}}:
Ai deception: A survey of examples, risks, and potential solutions.
Patterns
\textbf{5}(5)
(2024)
\end{botherref}
\endbibitem

%%% 4
\bibitem[\protect\citeauthoryear{Vinyals et~al.}{2019}]{vinyals2019grandmaster}
\begin{barticle}
\bauthor{\bsnm{Vinyals}, \binits{O.}},
\bauthor{\bsnm{Babuschkin}, \binits{I.}},
\bauthor{\bsnm{Czarnecki}, \binits{W.M.}},
\bauthor{\bsnm{Mathieu}, \binits{M.}},
\bauthor{\bsnm{Dudzik}, \binits{A.}},
\bauthor{\bsnm{Chung}, \binits{J.}},
\bauthor{\bsnm{Choi}, \binits{D.H.}},
\bauthor{\bsnm{Powell}, \binits{R.}},
\bauthor{\bsnm{Ewalds}, \binits{T.}},
\bauthor{\bsnm{Georgiev}, \binits{P.}}, \betal:
\batitle{Grandmaster level in starcraft ii using multi-agent reinforcement learning}.
\bjtitle{nature}
\bvolume{575}(\bissue{7782}),
\bfpage{350}--\blpage{354}
(\byear{2019})
\end{barticle}
\endbibitem

%%% 5
\bibitem[\protect\citeauthoryear{Brown and Sandholm}{2019}]{brown2019superhuman}
\begin{barticle}
\bauthor{\bsnm{Brown}, \binits{N.}},
\bauthor{\bsnm{Sandholm}, \binits{T.}}:
\batitle{Superhuman ai for multiplayer poker}.
\bjtitle{Science}
\bvolume{365}(\bissue{6456}),
\bfpage{885}--\blpage{890}
(\byear{2019})
\end{barticle}
\endbibitem

%%% 6
\bibitem[\protect\citeauthoryear{Christiano et~al.}{2017}]{christiano2017deep}
\begin{botherref}
\oauthor{\bsnm{Christiano}, \binits{P.F.}},
\oauthor{\bsnm{Leike}, \binits{J.}},
\oauthor{\bsnm{Brown}, \binits{T.}},
\oauthor{\bsnm{Martic}, \binits{M.}},
\oauthor{\bsnm{Legg}, \binits{S.}},
\oauthor{\bsnm{Amodei}, \binits{D.}}:
Deep reinforcement learning from human preferences.
Advances in neural information processing systems
\textbf{30}
(2017)
\end{botherref}
\endbibitem

%%% 7
\bibitem[\protect\citeauthoryear{Asimov}{2004}]{asimov2004robot}
\begin{bbook}
\bauthor{\bsnm{Asimov}, \binits{I.}}:
\bbtitle{I, Robot}
vol. \bseriesno{1}.
\bpublisher{Spectra},
\blocation{New York}
(\byear{2004})
\end{bbook}
\endbibitem

%%% 8
\bibitem[\protect\citeauthoryear{Schacter et~al.}{2012}]{schacter2012future}
\begin{barticle}
\bauthor{\bsnm{Schacter}, \binits{D.L.}},
\bauthor{\bsnm{Addis}, \binits{D.R.}},
\bauthor{\bsnm{Hassabis}, \binits{D.}},
\bauthor{\bsnm{Martin}, \binits{V.C.}},
\bauthor{\bsnm{Spreng}, \binits{R.N.}},
\bauthor{\bsnm{Szpunar}, \binits{K.K.}}:
\batitle{The future of memory: remembering, imagining, and the brain}.
\bjtitle{Neuron}
\bvolume{76}(\bissue{4}),
\bfpage{677}--\blpage{694}
(\byear{2012})
\end{barticle}
\endbibitem

%%% 9
\bibitem[\protect\citeauthoryear{D’Argembeau et~al.}{2008}]{d2008neural}
\begin{barticle}
\bauthor{\bsnm{D’Argembeau}, \binits{A.}},
\bauthor{\bsnm{Xue}, \binits{G.}},
\bauthor{\bsnm{Lu}, \binits{Z.-L.}},
\bauthor{\bsnm{Linden}, \binits{M.}},
\bauthor{\bsnm{Bechara}, \binits{A.}}:
\batitle{Neural correlates of envisioning emotional events in the near and far future}.
\bjtitle{Neuroimage}
\bvolume{40}(\bissue{1}),
\bfpage{398}--\blpage{407}
(\byear{2008})
\end{barticle}
\endbibitem

%%% 10
\bibitem[\protect\citeauthoryear{Hassabis and Maguire}{2009}]{hassabis2009construction}
\begin{barticle}
\bauthor{\bsnm{Hassabis}, \binits{D.}},
\bauthor{\bsnm{Maguire}, \binits{E.A.}}:
\batitle{The construction system of the brain}.
\bjtitle{Philosophical Transactions of the Royal Society B: Biological Sciences}
\bvolume{364}(\bissue{1521}),
\bfpage{1263}--\blpage{1271}
(\byear{2009})
\end{barticle}
\endbibitem

%%% 11
\bibitem[\protect\citeauthoryear{XU et~al.}{2015}]{xu2015imagining}
\begin{barticle}
\bauthor{\bsnm{XU}, \binits{X.}},
\bauthor{\bsnm{YU}, \binits{J.}},
\bauthor{\bsnm{LEI}, \binits{X.}}:
\batitle{Imagining the future: Cognitive processes and brain networks}.
\bjtitle{Advances in Psychological Science}
\bvolume{23}(\bissue{3}),
\bfpage{394}
(\byear{2015})
\end{barticle}
\endbibitem

%%% 12
\bibitem[\protect\citeauthoryear{Sebastian et~al.}{2012}]{sebastian2012neural}
\begin{barticle}
\bauthor{\bsnm{Sebastian}, \binits{C.L.}},
\bauthor{\bsnm{Fontaine}, \binits{N.M.}},
\bauthor{\bsnm{Bird}, \binits{G.}},
\bauthor{\bsnm{Blakemore}, \binits{S.-J.}},
\bauthor{\bsnm{De~Brito}, \binits{S.A.}},
\bauthor{\bsnm{McCrory}, \binits{E.J.}},
\bauthor{\bsnm{Viding}, \binits{E.}}:
\batitle{Neural processing associated with cognitive and affective theory of mind in adolescents and adults}.
\bjtitle{Social cognitive and affective neuroscience}
\bvolume{7}(\bissue{1}),
\bfpage{53}--\blpage{63}
(\byear{2012})
\end{barticle}
\endbibitem

%%% 13
\bibitem[\protect\citeauthoryear{Dennis et~al.}{2013}]{dennis2013cognitive}
\begin{barticle}
\bauthor{\bsnm{Dennis}, \binits{M.}},
\bauthor{\bsnm{Simic}, \binits{N.}},
\bauthor{\bsnm{Bigler}, \binits{E.D.}},
\bauthor{\bsnm{Abildskov}, \binits{T.}},
\bauthor{\bsnm{Agostino}, \binits{A.}},
\bauthor{\bsnm{Taylor}, \binits{H.G.}},
\bauthor{\bsnm{Rubin}, \binits{K.}},
\bauthor{\bsnm{Vannatta}, \binits{K.}},
\bauthor{\bsnm{Gerhardt}, \binits{C.A.}},
\bauthor{\bsnm{Stancin}, \binits{T.}}, \betal:
\batitle{Cognitive, affective, and conative theory of mind (tom) in children with traumatic brain injury}.
\bjtitle{Developmental cognitive neuroscience}
\bvolume{5},
\bfpage{25}--\blpage{39}
(\byear{2013})
\end{barticle}
\endbibitem

%%% 14
\bibitem[\protect\citeauthoryear{Zhang et~al.}{2018}]{zhang2018minimax}
\begin{bchapter}
\bauthor{\bsnm{Zhang}, \binits{S.}},
\bauthor{\bsnm{Durfee}, \binits{E.H.}},
\bauthor{\bsnm{Singh}, \binits{S.}}:
\bctitle{Minimax-regret querying on side effects for safe optimality in factored markov decision processes.}
In: \bbtitle{IJCAI},
pp. \bfpage{4867}--\blpage{4873}
(\byear{2018})
\end{bchapter}
\endbibitem

%%% 15
\bibitem[\protect\citeauthoryear{Irving et~al.}{2018}]{irving2018ai}
\begin{botherref}
\oauthor{\bsnm{Irving}, \binits{G.}},
\oauthor{\bsnm{Christiano}, \binits{P.}},
\oauthor{\bsnm{Amodei}, \binits{D.}}:
Ai safety via debate.
arXiv preprint arXiv:1805.00899
(2018)
\end{botherref}
\endbibitem

%%% 16
\bibitem[\protect\citeauthoryear{Armstrong and Levinstein}{2017}]{armstrong2017low}
\begin{botherref}
\oauthor{\bsnm{Armstrong}, \binits{S.}},
\oauthor{\bsnm{Levinstein}, \binits{B.}}:
Low impact artificial intelligences.
arXiv preprint arXiv:1705.10720
(2017)
\end{botherref}
\endbibitem

%%% 17
\bibitem[\protect\citeauthoryear{Krakovna et~al.}{2018}]{krakovna2018penalizing}
\begin{botherref}
\oauthor{\bsnm{Krakovna}, \binits{V.}},
\oauthor{\bsnm{Orseau}, \binits{L.}},
\oauthor{\bsnm{Kumar}, \binits{R.}},
\oauthor{\bsnm{Martic}, \binits{M.}},
\oauthor{\bsnm{Legg}, \binits{S.}}:
Penalizing side effects using stepwise relative reachability.
arXiv preprint arXiv:1806.01186
(2018)
\end{botherref}
\endbibitem

%%% 18
\bibitem[\protect\citeauthoryear{Turner et~al.}{2020a}]{turner2020avoiding}
\begin{barticle}
\bauthor{\bsnm{Turner}, \binits{A.}},
\bauthor{\bsnm{Ratzlaff}, \binits{N.}},
\bauthor{\bsnm{Tadepalli}, \binits{P.}}:
\batitle{Avoiding side effects in complex environments}.
\bjtitle{Advances in Neural Information Processing Systems}
\bvolume{33},
\bfpage{21406}--\blpage{21415}
(\byear{2020})
\end{barticle}
\endbibitem

%%% 19
\bibitem[\protect\citeauthoryear{Turner et~al.}{2020b}]{turner2020conservative}
\begin{bchapter}
\bauthor{\bsnm{Turner}, \binits{A.M.}},
\bauthor{\bsnm{Hadfield-Menell}, \binits{D.}},
\bauthor{\bsnm{Tadepalli}, \binits{P.}}:
\bctitle{Conservative agency via attainable utility preservation}.
In: \bbtitle{Proceedings of the AAAI/ACM Conference on AI, Ethics, and Society},
pp. \bfpage{385}--\blpage{391}
(\byear{2020})
\end{bchapter}
\endbibitem

%%% 20
\bibitem[\protect\citeauthoryear{Krakovna et~al.}{2020}]{krakovna2020avoiding}
\begin{barticle}
\bauthor{\bsnm{Krakovna}, \binits{V.}},
\bauthor{\bsnm{Orseau}, \binits{L.}},
\bauthor{\bsnm{Ngo}, \binits{R.}},
\bauthor{\bsnm{Martic}, \binits{M.}},
\bauthor{\bsnm{Legg}, \binits{S.}}:
\batitle{Avoiding side effects by considering future tasks}.
\bjtitle{Advances in Neural Information Processing Systems}
\bvolume{33},
\bfpage{19064}--\blpage{19074}
(\byear{2020})
\end{barticle}
\endbibitem

%%% 21
\bibitem[\protect\citeauthoryear{Bussmann et~al.}{2019}]{bussmann2019towards}
\begin{bchapter}
\bauthor{\bsnm{Bussmann}, \binits{B.}},
\bauthor{\bsnm{Heinerman}, \binits{J.}},
\bauthor{\bsnm{Lehman}, \binits{J.}}:
\bctitle{Towards empathic deep q-learning}.
In: \bbtitle{2019 Workshop on Artificial Intelligence Safety, AISafety 2019},
pp. \bfpage{1}--\blpage{7}
(\byear{2019}).
\bcomment{CEUR-WS. org}
\end{bchapter}
\endbibitem

%%% 22
\bibitem[\protect\citeauthoryear{Senadeera et~al.}{2022}]{senadeera2022sympathy}
\begin{bchapter}
\bauthor{\bsnm{Senadeera}, \binits{M.}},
\bauthor{\bsnm{Karimpanal}, \binits{T.G.}},
\bauthor{\bsnm{Gupta}, \binits{S.}},
\bauthor{\bsnm{Rana}, \binits{S.}}:
\bctitle{Sympathy-based reinforcement learning agents}.
In: \bbtitle{Proceedings of the 21st International Conference on Autonomous Agents and Multiagent Systems},
pp. \bfpage{1164}--\blpage{1172}
(\byear{2022})
\end{bchapter}
\endbibitem

%%% 23
\bibitem[\protect\citeauthoryear{Alizadeh~Alamdari et~al.}{2022}]{alizadeh2022considerate}
\begin{bchapter}
\bauthor{\bsnm{Alizadeh~Alamdari}, \binits{P.}},
\bauthor{\bsnm{Klassen}, \binits{T.Q.}},
\bauthor{\bsnm{Toro~Icarte}, \binits{R.}},
\bauthor{\bsnm{McIlraith}, \binits{S.A.}}:
\bctitle{Be considerate: Avoiding negative side effects in reinforcement learning}.
In: \bbtitle{Proceedings of the 21st International Conference on Autonomous Agents and Multiagent Systems},
pp. \bfpage{18}--\blpage{26}
(\byear{2022})
\end{bchapter}
\endbibitem

%%% 24
\bibitem[\protect\citeauthoryear{Klassen et~al.}{2022}]{klassen2022epistemic}
\begin{bchapter}
\bauthor{\bsnm{Klassen}, \binits{T.Q.}},
\bauthor{\bsnm{Alamdari}, \binits{P.A.}},
\bauthor{\bsnm{McIlraith}, \binits{S.A.}}:
\bctitle{Epistemic side effects \& avoiding them (sometimes)}.
In: \bbtitle{NeurIPS ML Safety Workshop}
(\byear{2022})
\end{bchapter}
\endbibitem

%%% 25
\bibitem[\protect\citeauthoryear{Klassen et~al.}{2023}]{klassen2023epistemic}
\begin{bchapter}
\bauthor{\bsnm{Klassen}, \binits{T.Q.}},
\bauthor{\bsnm{Alamdari}, \binits{P.A.}},
\bauthor{\bsnm{McIlraith}, \binits{S.A.}}:
\bctitle{Epistemic side effects: An ai safety problem}.
In: \bbtitle{Proceedings of the 2023 International Conference on Autonomous Agents and Multiagent Systems},
pp. \bfpage{1797}--\blpage{1801}
(\byear{2023})
\end{bchapter}
\endbibitem

%%% 26
\bibitem[\protect\citeauthoryear{Feng et~al.}{2022}]{feng2022brain}
\begin{barticle}
\bauthor{\bsnm{Feng}, \binits{H.}},
\bauthor{\bsnm{Zeng}, \binits{Y.}},
\bauthor{\bsnm{Lu}, \binits{E.}}:
\batitle{Brain-inspired affective empathy computational model and its application on altruistic rescue task}.
\bjtitle{Frontiers in Computational Neuroscience}
\bvolume{16},
\bfpage{784967}
(\byear{2022})
\end{barticle}
\endbibitem

%%% 27
\bibitem[\protect\citeauthoryear{Zhao et~al.}{2024}]{zhao2024building}
\begin{botherref}
\oauthor{\bsnm{Zhao}, \binits{F.}},
\oauthor{\bsnm{Feng}, \binits{H.}},
\oauthor{\bsnm{Tong}, \binits{H.}},
\oauthor{\bsnm{Han}, \binits{Z.}},
\oauthor{\bsnm{Lu}, \binits{E.}},
\oauthor{\bsnm{Sun}, \binits{Y.}},
\oauthor{\bsnm{Zeng}, \binits{Y.}}:
Building altruistic and moral ai agent with brain-inspired affective empathy mechanisms.
arXiv preprint arXiv:2410.21882
(2024)
\end{botherref}
\endbibitem

%%% 28
\bibitem[\protect\citeauthoryear{Zhao et~al.}{2022}]{zhao2022brain}
\begin{barticle}
\bauthor{\bsnm{Zhao}, \binits{Z.}},
\bauthor{\bsnm{Lu}, \binits{E.}},
\bauthor{\bsnm{Zhao}, \binits{F.}},
\bauthor{\bsnm{Zeng}, \binits{Y.}},
\bauthor{\bsnm{Zhao}, \binits{Y.}}:
\batitle{A brain-inspired theory of mind spiking neural network for reducing safety risks of other agents}.
\bjtitle{Frontiers in neuroscience}
\bvolume{16},
\bfpage{753900}
(\byear{2022})
\end{barticle}
\endbibitem

%%% 29
\bibitem[\protect\citeauthoryear{Zhao et~al.}{2023}]{zhao2023brain}
\begin{botherref}
\oauthor{\bsnm{Zhao}, \binits{Z.}},
\oauthor{\bsnm{Zhao}, \binits{F.}},
\oauthor{\bsnm{Zhao}, \binits{Y.}},
\oauthor{\bsnm{Zeng}, \binits{Y.}},
\oauthor{\bsnm{Sun}, \binits{Y.}}:
A brain-inspired theory of mind spiking neural network improves multi-agent cooperation and competition.
Patterns
\textbf{4}(8)
(2023)
\end{botherref}
\endbibitem

%%% 30
\bibitem[\protect\citeauthoryear{Ray et~al.}{2019}]{ray2019benchmarking}
\begin{barticle}
\bauthor{\bsnm{Ray}, \binits{A.}},
\bauthor{\bsnm{Achiam}, \binits{J.}},
\bauthor{\bsnm{Amodei}, \binits{D.}}:
\batitle{Benchmarking safe exploration in deep reinforcement learning}.
\bjtitle{arXiv preprint arXiv:1910.01708}
\bvolume{7}(\bissue{1}),
\bfpage{2}
(\byear{2019})
\end{barticle}
\endbibitem

%%% 31
\bibitem[\protect\citeauthoryear{Ji et~al.}{2023}]{ji2023safety}
\begin{botherref}
\oauthor{\bsnm{Ji}, \binits{J.}},
\oauthor{\bsnm{Zhang}, \binits{B.}},
\oauthor{\bsnm{Zhou}, \binits{J.}},
\oauthor{\bsnm{Pan}, \binits{X.}},
\oauthor{\bsnm{Huang}, \binits{W.}},
\oauthor{\bsnm{Sun}, \binits{R.}},
\oauthor{\bsnm{Geng}, \binits{Y.}},
\oauthor{\bsnm{Zhong}, \binits{Y.}},
\oauthor{\bsnm{Dai}, \binits{J.}},
\oauthor{\bsnm{Yang}, \binits{Y.}}:
Safety gymnasium: A unified safe reinforcement learning benchmark.
Advances in Neural Information Processing Systems
\textbf{36}
(2023)
\end{botherref}
\endbibitem

%%% 32
\bibitem[\protect\citeauthoryear{Wainwright and Eckersley}{2019}]{wainwright2019safelife}
\begin{botherref}
\oauthor{\bsnm{Wainwright}, \binits{C.L.}},
\oauthor{\bsnm{Eckersley}, \binits{P.}}:
Safelife 1.0: Exploring side effects in complex environments.
arXiv preprint arXiv:1912.01217
(2019)
\end{botherref}
\endbibitem

%%% 33
\bibitem[\protect\citeauthoryear{Dulac-Arnold et~al.}{2020}]{dulac2020empirical}
\begin{botherref}
\oauthor{\bsnm{Dulac-Arnold}, \binits{G.}},
\oauthor{\bsnm{Levine}, \binits{N.}},
\oauthor{\bsnm{Mankowitz}, \binits{D.J.}},
\oauthor{\bsnm{Li}, \binits{J.}},
\oauthor{\bsnm{Paduraru}, \binits{C.}},
\oauthor{\bsnm{Gowal}, \binits{S.}},
\oauthor{\bsnm{Hester}, \binits{T.}}:
An empirical investigation of the challenges of real-world reinforcement learning.
arXiv preprint arXiv:2003.11881
(2020)
\end{botherref}
\endbibitem

%%% 34
\bibitem[\protect\citeauthoryear{Mnih et~al.}{2015}]{mnih2015human}
\begin{barticle}
\bauthor{\bsnm{Mnih}, \binits{V.}},
\bauthor{\bsnm{Kavukcuoglu}, \binits{K.}},
\bauthor{\bsnm{Silver}, \binits{D.}},
\bauthor{\bsnm{Rusu}, \binits{A.A.}},
\bauthor{\bsnm{Veness}, \binits{J.}},
\bauthor{\bsnm{Bellemare}, \binits{M.G.}},
\bauthor{\bsnm{Graves}, \binits{A.}},
\bauthor{\bsnm{Riedmiller}, \binits{M.}},
\bauthor{\bsnm{Fidjeland}, \binits{A.K.}},
\bauthor{\bsnm{Ostrovski}, \binits{G.}}, \betal:
\batitle{Human-level control through deep reinforcement learning}.
\bjtitle{nature}
\bvolume{518}(\bissue{7540}),
\bfpage{529}--\blpage{533}
(\byear{2015})
\end{barticle}
\endbibitem

%%% 35
\bibitem[\protect\citeauthoryear{Maass}{1997}]{maass1997networks}
\begin{barticle}
\bauthor{\bsnm{Maass}, \binits{W.}}:
\batitle{Networks of spiking neurons: the third generation of neural network models}.
\bjtitle{Neural networks}
\bvolume{10}(\bissue{9}),
\bfpage{1659}--\blpage{1671}
(\byear{1997})
\end{barticle}
\endbibitem

%%% 36
\bibitem[\protect\citeauthoryear{Nair and Hinton}{2010}]{nair2010rectified}
\begin{bchapter}
\bauthor{\bsnm{Nair}, \binits{V.}},
\bauthor{\bsnm{Hinton}, \binits{G.E.}}:
\bctitle{Rectified linear units improve restricted boltzmann machines}.
In: \bbtitle{Proceedings of the 27th International Conference on Machine Learning (ICML-10)},
pp. \bfpage{807}--\blpage{814}
(\byear{2010})
\end{bchapter}
\endbibitem

%%% 37
\bibitem[\protect\citeauthoryear{Dayan and Abbott}{2005}]{dayan2005theoretical}
\begin{bbook}
\bauthor{\bsnm{Dayan}, \binits{P.}},
\bauthor{\bsnm{Abbott}, \binits{L.F.}}:
\bbtitle{Theoretical Neuroscience: Computational and Mathematical Modeling of Neural Systems}.
\bpublisher{MIT press}, \blocation{???}
(\byear{2005})
\end{bbook}
\endbibitem

%%% 38
\bibitem[\protect\citeauthoryear{Zeng et~al.}{2023}]{zeng2023braincog}
\begin{botherref}
\oauthor{\bsnm{Zeng}, \binits{Y.}},
\oauthor{\bsnm{Zhao}, \binits{D.}},
\oauthor{\bsnm{Zhao}, \binits{F.}},
\oauthor{\bsnm{Shen}, \binits{G.}},
\oauthor{\bsnm{Dong}, \binits{Y.}},
\oauthor{\bsnm{Lu}, \binits{E.}},
\oauthor{\bsnm{Zhang}, \binits{Q.}},
\oauthor{\bsnm{Sun}, \binits{Y.}},
\oauthor{\bsnm{Liang}, \binits{Q.}},
\oauthor{\bsnm{Zhao}, \binits{Y.}}, et al.:
Braincog: A spiking neural network based, brain-inspired cognitive intelligence engine for brain-inspired ai and brain simulation.
Patterns
\textbf{4}(8)
(2023)
\end{botherref}
\endbibitem

%%% 39
\bibitem[\protect\citeauthoryear{Shen et~al.}{2022}]{shen2022backpropagation}
\begin{botherref}
\oauthor{\bsnm{Shen}, \binits{G.}},
\oauthor{\bsnm{Zhao}, \binits{D.}},
\oauthor{\bsnm{Zeng}, \binits{Y.}}:
Backpropagation with biologically plausible spatiotemporal adjustment for training deep spiking neural networks.
Patterns
\textbf{3}(6)
(2022)
\end{botherref}
\endbibitem

%%% 40
\bibitem[\protect\citeauthoryear{Wu et~al.}{2024}]{wu2024motive}
\begin{botherref}
\oauthor{\bsnm{Wu}, \binits{X.}},
\oauthor{\bsnm{Ren}, \binits{X.}},
\oauthor{\bsnm{Liu}, \binits{C.}},
\oauthor{\bsnm{Zhang}, \binits{H.}}:
The motive cocktail in altruistic behaviors.
Nature Computational Science,
1--18
(2024)
\end{botherref}
\endbibitem

%%% 41
\bibitem[\protect\citeauthoryear{Jin et~al.}{2024}]{jin2024observing}
\begin{barticle}
\bauthor{\bsnm{Jin}, \binits{K.}},
\bauthor{\bsnm{Wu}, \binits{J.}},
\bauthor{\bsnm{Zhang}, \binits{R.}},
\bauthor{\bsnm{Zhang}, \binits{S.}},
\bauthor{\bsnm{Wu}, \binits{X.}},
\bauthor{\bsnm{Wu}, \binits{T.}},
\bauthor{\bsnm{Gu}, \binits{R.}},
\bauthor{\bsnm{Liu}, \binits{C.}}:
\batitle{Observing heroic behavior and its influencing factors in immersive virtual environments}.
\bjtitle{Proceedings of the National Academy of Sciences}
\bvolume{121}(\bissue{17}),
\bfpage{2314590121}
(\byear{2024})
\end{barticle}
\endbibitem

%%% 42
\bibitem[\protect\citeauthoryear{Rueckauer et~al.}{2017}]{rueckauer2017conversion}
\begin{barticle}
\bauthor{\bsnm{Rueckauer}, \binits{B.}},
\bauthor{\bsnm{Lungu}, \binits{I.-A.}},
\bauthor{\bsnm{Hu}, \binits{Y.}},
\bauthor{\bsnm{Pfeiffer}, \binits{M.}},
\bauthor{\bsnm{Liu}, \binits{S.-C.}}:
\batitle{Conversion of continuous-valued deep networks to efficient event-driven networks for image classification}.
\bjtitle{Frontiers in neuroscience}
\bvolume{11},
\bfpage{682}
(\byear{2017})
\end{barticle}
\endbibitem

\end{thebibliography}

\end{document}